\documentclass[conference]{IEEEtran}
\IEEEoverridecommandlockouts

\usepackage{graphicx}
\usepackage{textcomp}
\usepackage{enumitem}

\usepackage{tabularx}
 \usepackage{textcomp}

\usepackage{booktabs}

\usepackage{color}
\usepackage{url}
\usepackage{algorithm,algpseudocode}
\usepackage{enumitem}
\usepackage{multirow}
\usepackage{dirtytalk}
\usepackage{balance}

\usepackage{color}

\usepackage{pifont}
\usepackage{float}
\usepackage{placeins}

\usepackage{graphicx}
\usepackage{eqnarray}
\usepackage{rotating}
\usepackage{color}
\usepackage{booktabs}

\usepackage{colortbl}

\usepackage{framed}

\usepackage{tikz}
\usepackage{textcomp}
\newcommand\copyrighttext{%
  \footnotesize \textcopyright 2026 IEEE. Personal use of this material is permitted.  Permission from IEEE must be obtained for all other uses, in any current or future media, including reprinting/republishing this material for advertising or promotional purposes, creating new collective works, for resale or redistribution to servers or lists, or reuse of any copyrighted component of this work in other works.
  Presented at CBDCOM 2025.}
\newcommand{\copyrightnotice}{%
\begin{tikzpicture}[remember picture,overlay]
\node[anchor=south,yshift=10pt] at (current page.south) {\fbox{\parbox{\dimexpr\textwidth-\fboxsep-\fboxrule\relax}{\copyrighttext}}};
\end{tikzpicture}%
}

\newcommand{\remove}[1]{ }
\usepackage{cite}
\usepackage{amsmath,amssymb,amsfonts}
\usepackage{graphicx}
\usepackage{textcomp}
\usepackage{xcolor}
\def\BibTeX{{\rm B\kern-.05em{\sc i\kern-.025em b}\kern-.08em
    T\kern-.1667em\lower.7ex\hbox{E}\kern-.125emX}}
\begin{document}

\title{Self-adaptive Multi-Access Edge Architectures: A Robotics Case}

\author{
\IEEEauthorblockN{Mahyar T. Moghaddam}
\IEEEauthorblockA{%
\textit{University of Southern Denmark}\\
Odense, Denmark\\
mtmo@mmmi.sdu.dk}

\and
\IEEEauthorblockN{Joakim V. Leed}
\IEEEauthorblockA{%
\textit{University of Southern Denmark}\\
Odense, Denmark\\
jolee18@student.sdu.dk}

\and
\IEEEauthorblockN{Anders Q. Frandsen}
\IEEEauthorblockA{%
\textit{University of Southern Denmark}\\
Odense, Denmark\\
anfra22@student.sdu.dk}
}



\maketitle

\copyrightnotice

\begin{abstract}
The growth of compute-intensive AI tasks highlights the need to mitigate the processing costs and improve performance and energy efficiency. This necessitates the integration of intelligent agents as architectural adaptation supervisors tasked with adaptive scaling of the infrastructure and efficient offloading of computation within the continuum. This paper presents a self-adaptation approach for an efficient computing system of a mixed human-robot environment. The computation task is associated with a Neural Network algorithm that leverages sensory data to predict human mobility behaviors, to enhance mobile robots' proactive path planning, and ensure human safety. To streamline neural network processing, we built a distributed edge offloading system with heterogeneous processing units, orchestrated by Kubernetes. By monitoring response times and power consumption, the MAPE-K-based adaptation supervisor makes informed decisions on scaling and offloading. Results show notable improvements in service quality over traditional setups, demonstrating the effectiveness of the proposed approach for AI-driven systems.
\end{abstract}


\section{Introduction}

The rise of compute-intensive artificial intelligence (AI) workloads has created a significant challenge for smart systems (such as robotics \cite{khoramshahi2019dynamical}) working in environments shared with humans: how to achieve low-latency responses and energy efficiency \cite{wang2024load} while ensuring safety and productivity. In factory and logistics scenarios, mobile robots rely on predictive AI services to anticipate human trajectories \cite{rudenko2020human}, avoid collisions, and plan routes with hard real-time requirements.
Traditional cloud‑centric deployments \cite{tourchi2023architecting} \cite{tourchi2023energy} suffer from network delays and resource contention \cite{al2024real}, whereas isolated edge devices often lack sufficient computational power and become overloaded as workloads fluctuate \cite{mansouri2021review}. Emerging multi‑access edge computing {\em (MEC)} aims to address these limitations by deploying heterogeneous computing resources (small embedded devices, mini‑PCs, and servers) at the edge, thereby reducing latency and energy consumption. However, exploiting MEC effectively requires adaptive architectural mechanisms \cite{weyns2020introduction} that can decide when to offload computation to different nodes, scale resources on demand, and balance quality‑of‑service attributes.

Existing solutions often rely on static or rule-based task allocation, lacking flexibility for dynamic environments. While some use heuristic or model-driven offloading \cite{liang2023model}, they rarely integrate real-time feedback for AI tasks. This work proposes a MAPE-K-based self-adaptive architecture (Monitor, Analyze, Plan, Execute, under a knowledge base) \cite{moghaddam2020ias} \cite{moghaddam2022hierarchical} that combines real-time monitoring, modeling, and Kubernetes orchestration to optimize offloading and resource use.
The proposed architecture uses feedback-driven adaptation to optimize AI task distribution and resource use. The AI task is an LSTM network for predicting human mobility, enabling proactive, conflict-free human-robot collaboration. Tasks are distributed across heterogeneous nodes (Raspberry Pi and ASUS PN53) managed via Kubernetes. A custom Rust-based metrics server provides real-time data on CPU load, latency, and power usage. Pod- and service-level load balancing dynamically adjusts task allocation based on these metrics.
Evaluation took place in a real assembly line where ultra-wideband (UWB) localization was fed to LSTM, thus the system predicted human movement to optimize robot routing. Robots were assigned missions under varying workloads, with compute-heavy tasks offloaded across edge nodes. Response time, energy use, and resource utilization were measured to assess adaptability and efficiency.

The research focuses on addressing the following {\bf \em questions}:
{\em RQ1: Impact of adaptive offloading.} How does an adaptive offloading strategy affect response time and energy consumption compared to static offloading? 
{\em RQ2: Granularity of load balancing.} How does pod‑level load balancing compare with service‑level load balancing? 
{\em RQ3.)} How does dynamic autoscaling influence the trade‑off between response time and energy consumption? 


The main contributions of this work are {\em i)} providing a literature review on offloading techniques for architecture adaptation (Section II), {\em ii)} presenting a self-adaptive architecture approach that dynamically scales resources and optimizes AI task offloading using real-time feedback (Section III), {\em iii)}  presenting a real case and detailing its implementation (Section IV)  followed by empirical evaluation of the approach's efficiency (Section V).
 The paper also provides lessons learned in Section VI and concludes with a summary of findings in Section VII.


\section{Literature Review}


\subsection{Mixed Human-Robot Environments}

Mixed human-robot environments, especially in manufacturing, require advanced systems to process diverse sensor data such as vision, UWB, IMUs, and environmental inputs for safe and efficient collaboration \cite{proia2021control, bonci2021human}. These systems demand low-latency decision-making, real-time streaming, and predictive analytics.
AI frameworks, particularly those using RNNs like LSTMs, are widely used to predict human motion for tasks like collision avoidance and adaptive scheduling \cite{quan2021holistic}. Hybrid architectures combining edge and cloud computing are commonly adopted to balance low-latency needs with heavy processing demands \cite{shukur2020state, venu2022secure}.
Cognitive architectures (e.g., SOAR, ACT-R) enhanced with reinforcement learning further support robot adaptability in collaborative tasks \cite{ren2023human, laird2022introduction}. However, many existing solutions lack tight integration between predictive AI and real-time task adaptation, especially in dynamically shared human–robot spaces.

\subsection{Adaptation Supervision Techniques}

To meet quality objectives in dynamic environments, offloading from sources like mobile phones \cite{MAUI}, vehicles \cite{GA-Vehicular}, or IoT devices \cite{fuzzy-offload} \cite{muccini2018iot} should be handled systematically. 
{\bf \em Rule-based} approaches (e.g., always-offload, random-offload) apply static policies without optimization. They are often used as baselines or in comparative studies \cite{Deep-compression, escove-ismail2021, Minority-game}.
{\bf \em Model-based} methods represent system behavior and enable predictive decision-making. Model-driven engineering approaches describe systems at design and runtime \cite{de2021survey}. For instance, \cite{liang2023model} builds analytic models to optimize edge inference performance, while \cite{leppanen2020service} applies model-based development for MEC service design. Architectural models support feedback loops like MAPE-K for self-adaptation \cite{weyns2018applying}, as seen in context-aware systems \cite{nakahara2018context}.

{\bf \em Optimization} techniques range from fuzzy logic to meta-heuristics:
{\em i)} {Fuzzy-based offloading policies handle imprecise classifications (e.g., data-intensive tasks) \cite{fuzzy-offload}.
{\em ii)} Evolutionary algorithms (e.g., Genetic Algorithm) solve NP-hard offloading problems efficiently \cite{Meta-heustic-for-NP-hard, abbas2021meta}.
{\em iii)} Lyapunov optimization ensures system stability while improving performance, particularly in wireless MEC \cite{Power-latency-MEC}.
{\em iv)} Markov Decision Processes (MDP) optimize task placement based on probabilistic state transitions \cite{Time-optimization-MDP}.

{\bf \em Game-theoretic} approaches use agent-based coordination to manage offloading in decentralized or cooperative settings \cite{Minority-game, Collaborative-MEC}.
{\bf \em Heuristic} methods like regression-based estimators can be integrated into scheduling algorithms to approximate energy or delay costs under constraints \cite{escove-ismail2021, GA-Vehicular}.
{\bf \em Reinforcement learning (RL)}, including Q-learning, supports adaptive offloading by learning optimal actions from experience. RL models define rewards based on metrics like latency, energy, and QoE, using temporal difference learning to refine decisions over time \cite{UAV-Q-learning}.

While these methods show promise, most lack integration with real-time, high-frequency metrics and often do not scale well with heterogeneous infrastructure. The need remains for adaptive architectures that respond dynamically to workload variability in realistic environments.

\subsection{Quality Attributes}

In computation-intensive systems, key quality attributes (performance, energy efficiency, and scalability) are tightly linked to offloading and resource management strategies, directly influencing system responsiveness and sustainability.

{\bf \em Performance} is critical in real-time applications like robot task scheduling and mobility prediction \cite{mondal2024optimal}. Kubernetes-based dynamic offloading can improve latency and throughput \cite{nguyen2022load}, while static configurations often struggle under variable workloads, leading to bottlenecks and resource underuse \cite{dong2024task}.

{\bf \em Energy Efficiency} is increasingly important in heterogeneous edge systems. While lightweight nodes like Raspberry Pis conserve power, they lack the capacity for compute-heavy tasks. In contrast, GPUs offer speed but at higher energy costs \cite{neill2020overview}. Reinforcement learning has been applied to energy-aware offloading \cite{rahimi2022cloud}, but real-time energy profiling remains underexplored and essential for continuous operation in dynamic settings.

{\bf \em Scalability} ensures system stability under varying load conditions. Horizontal scaling helps manage load spikes but can introduce latency during provisioning \cite{tari2024auto}. Predictive scaling using AI forecasts offers reduced delays \cite{ramamoorthi2024ai}, and federated edge-cloud frameworks improve task distribution and reliability across resource tiers \cite{kar2023offloading}.

This work extends prior research by integrating real-time feedback, adaptive scheduling, and predictive scaling in a unified architecture. By combining pod-level load balancing with dynamic resource allocation, the system meets the performance, energy, and scalability demands of modern mixed human-robot environments.

\section{Approach}

We adopt a MAPE-K self-adaptive framework \cite{muccini2018self} to optimize task offloading and resource management.
As shown in Figure \ref{appr}, we compare category-level and pod-level adaptive offloading strategies. The goal is to ensure high performance and energy efficiency under dynamic workloads by leveraging Kubernetes orchestration, real-time monitoring, and intelligent decision-making.

\subsection{Category-Level Adaptation} 

In this strategy, the Plan phase assigns tasks at the granularity of node categories (e.g., “small” Raspberry Pi or “medium” PN53 servers). The managing system directs each incoming task to a Kubernetes Service corresponding to a category. Kubernetes’ built-in service-level load balancer (kube-proxy with IPTables in K3s) then forwards the task to one of the pods in that category, typically in a round-robin fashion. This approach aligns with a traditional load-balancing schema: the adaptation logic decides which class of nodes should handle the task, but leaves the selection of a specific pod to Kubernetes. Within the MAPE-K loop, {\em Monitor} collects category-level metrics (e.g., average response time per service and each node’s CPU load), and {\em Analyze} identifies imbalances between categories (such as the small-node service becoming a bottleneck). {\em Plan} might adjust the fraction of tasks sent to each category (for example, to offload more tasks to medium nodes when small nodes are overloaded) and can {\em Execute} trigger scaling of entire deployments if needed. The adaptation strategy is shown in Algorithm \ref{alg1} where $w_1$ and $w_2$ are weighting coefficients tuned empirically and $S[c]$ is the score of category c.

\begin{algorithm}[t]
\caption{Category-Level Adaptation}\label{alg:category}
\begin{scriptsize}
\begin{algorithmic}[1]
\Require Categories $C = \{\text{small}, \text{medium}\}$; Task $T$; 
CPU usage $\text{CPU}[c]$ and response times $\text{RT}[c]$ per category
\Procedure{AssignCategory}{$T, C$}
    \State Compute cost score for each category:
    \ForAll{$c \in C$}
        \State $S[c] \gets w_1 \cdot \text{RT}[c] + w_2 \cdot \text{CPU}[c]$
    \EndFor
    \State $c^* \gets \arg\min_{c} S[c]$
    \State Dispatch $T$ to Kubernetes service representing $c^*$
\EndProcedure
\end{algorithmic}
\end{scriptsize}
\label{alg1}
\end{algorithm}


\subsection{Pod-Level Adaptation}

This strategy shifts the decision-making to a finer granularity. Here, the managing system’s Plan phase assigns tasks directly to individual pods based on real-time performance metrics. In the MAPE-K loop, {\em Monitor} (via a custom agent) gathers detailed metrics for each pod (e.g., per-pod response times, CPU utilization, and energy consumption). {\em Analyze} uses this data to detect overloaded pods or underutilized resources across the cluster. Based on this analysis, {\em Plan} updates a Service Profile, a data structure mapping each pod (identified by its node and service) to an offloading probability or priority. This profile essentially encodes the decision policy: for example, if a certain PN53 pod exhibits the lowest latency, its probability of receiving the next task is raised. The offloading agent (in {\em Execute}) then uses this profile to route each incoming task directly to a specific pod’s IP (bypassing the Kubernetes service round-robin) via asynchronous HTTP requests as shown in Algorithm \ref{alg2}.

\begin{algorithm}[h]
\caption{Pod-Level Adaptation}\label{alg:pod}
\begin{scriptsize}
\begin{algorithmic}[1]
\Require Pods $P$; Task $T$; Metrics $\text{RT}[p]$, $\text{CPU}[p]$, $\text{Energy}[p]$; Weights $w_1$, $w_2$, $w_3$
\Procedure{AssignPod}{$T, P$}
    \State Compute cost score for each pod:
    \ForAll{$p \in P$}
        \State $S[p] \gets w_1 \cdot \text{RT}[p] + w_2 \cdot \text{CPU}[p] + w_3 \cdot \text{Energy}[p]$
    \EndFor
    \State Normalize $S$ to probabilities $P_{\text{norm}}$
    \State $p^* \gets \text{sample\_from}(P_{\text{norm}})$
    \State Dispatch $T$ directly to IP($p^*$) via HTTP
\EndProcedure
\end{algorithmic}
\end{scriptsize}
\label{alg2}
\end{algorithm}

Figure \ref{appr} illustrates these strategies on a MAPE-K-based architecture.
The top of the figure illustrates the adaptation mechanisms in the Kubernetes-managed cluster, where task distribution and resource allocation are guided by real-time metrics like response times, utilization, and energy consumption. These insights update the Service Profile, enabling the system to compute optimal offloading probabilities. Tasks are then executed on target pods or services using pod- and service-level load balancing strategies.

\begin{figure}[ht]
    \centering
    \includegraphics[width=1\columnwidth]{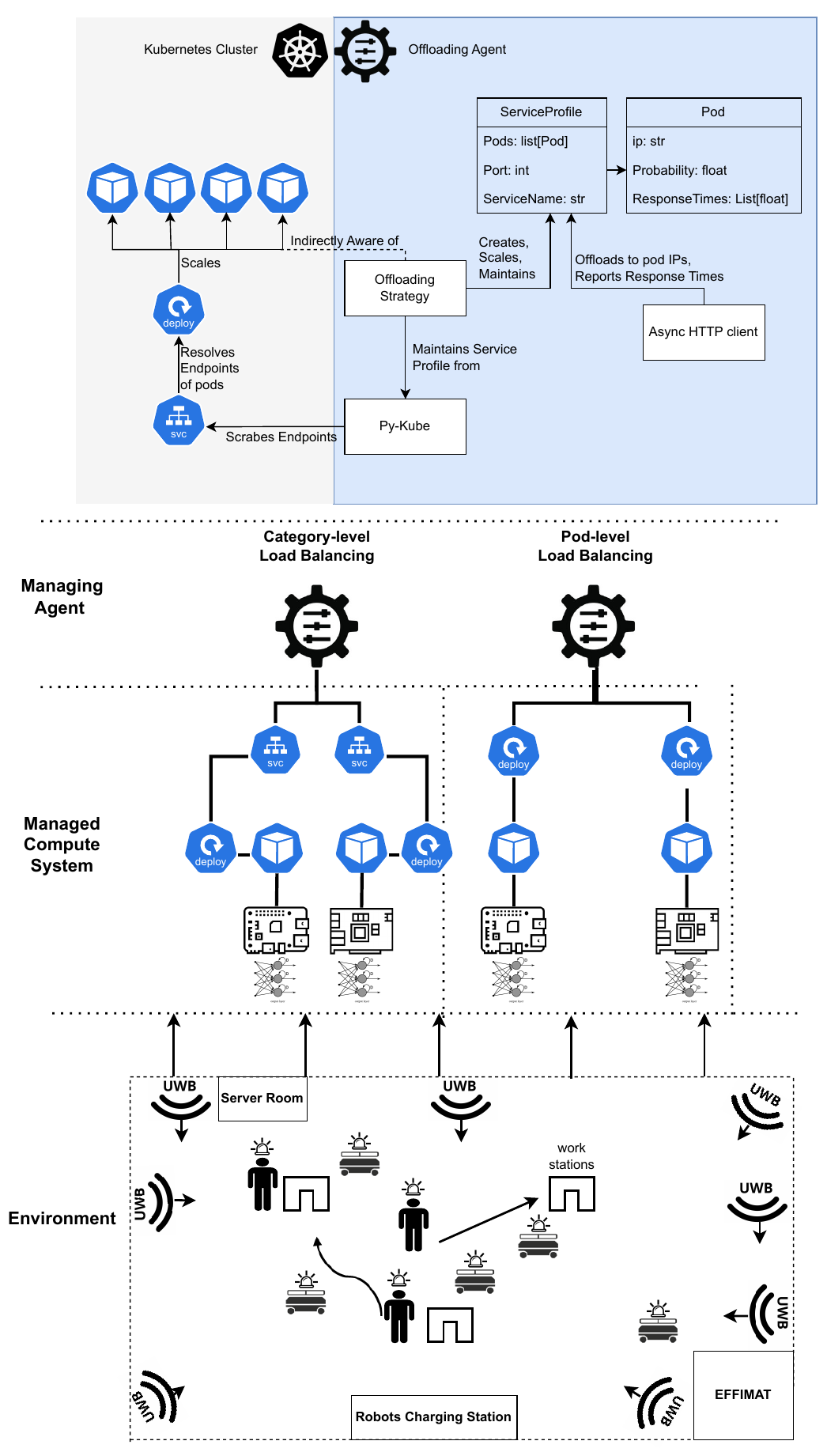}
    \caption{The approach showing the interaction of the environment and managed and managing systems.}
    \label{appr}
\end{figure}

\section{Real Case}



The motivating case is a mixed assembly line (see the bottom of Figure \ref{appr}) where mobile robots coordinate with human operators and stationary machines.
A Ubisense UWB-localization infrastructure (consisting of 8 sensors) was installed to sense and record the position of people or moving objects over time. The system is an advanced real-time location system (RTLS) that provides 3D tracking of objects within an environment.  The system utilizes a combination of tags and sensors to pinpoint and track objects' positions in real-time.
Getting input from UWB, a Long Short-term
Memory Recurrent Neural Network, LSTM-RNN, implemented in Python with Keras and Tensorflow was employed to predict human movement trajectories. LSTM memory cell architecture allows handling long-term
dependencies with complex cell structures. While it does spend more time on the training phase, it performs noticeably better than traditional RNN.
The aim is to retrain and predict new data online.
The LSTM model architecture included configurations optimized for predicting 5-second mobility horizons.
It achieved over 90\% prediction accuracy during training and was deployed using Docker for live predictions during experiments.
Predictions were transmitted via ROS wrappers to the Fleet Manager, enabling real-time adjustments to robot trajectories and collision avoidance strategies.



{\bf Hardware and infrastructure.} The hardware infrastructure included 19 Ubisense tags paired with UWB sensors, which provided real-time position tracking for human and robotic agents. The sensor data was streamed to an MQTT broker at one-second intervals. The computational resources were distributed across a server rack that housed three PN53 servers with AMD Ryzen 5 7535HS CPUs and six Raspberry Pi 4b devices. Three of the Raspberry Pis were designated as Kubernetes control planes, while the remaining nodes formed the cluster's heterogeneous processing units, categorized into small (Raspberry Pi) and medium (PN53) nodes. The offloading target (LSTM), compiled to run on ARM64 architecture using Docker and QEMU for compatibility with the Raspberry Pi devices.
An HMC electricity analytics device from RSInstruments is installed to grant us real-time power consumption of the rack (Figure \ref{rack}).

{\bf Software components.} The software stack supporting the system included the offloading agent, which subscribed to the MQTT broker to retrieve sensor data and forwarded it to a REST endpoint for processing. The Kubernetes cluster managed the deployment and operation of pods and services, facilitating dynamic task distribution through pod-level and service-level load balancing.
The response time is measured as the difference between the time a request is offloaded from the offloading agent and the time it is returned from the mobility behavior neural network.
Resource utilization data, including CPU and memory usage, was collected in real-time using a custom-built Rust metrics server, which supplemented the default Kubernetes metrics server by providing historic CPU and memory data at 200-millisecond intervals. The system's behavior was further analyzed using profiling tools, including cProfiler for function-level profiling of the offloading application and the Perf tool for kernel-level and userspace process monitoring on all worker nodes.

{\bf Experiment workflow.} 
The robots operated in an environment occupied by 5 human operators. The participants also received their missions through a location-aware mobile interface and moved between the workstations. Along with the robot data, the robots and the human participants were equipped with a UWB transmitter to track their locations.
The run-time and predicted data (in a horizon of 5 seconds) on agents' mobility were transferred to the fleet through a ROS wrapper for further routing planning.
The heterogeneous server was configured with three control planes consisting of Raspberry Pi, and the remaining servers were designated worker nodes. The offloading agent would ensure that service and deployment were created for each node category and that every pod of the deployment was scaled to a minimum of five pods.

All pods were reset to ensure CPU utilization was below 100m, establishing a clean baseline. Task configurations were defined, including experiment duration (15 minutes), the number of iterations (1), offloading coroutines (20), and specific offloading strategies. During each experiment, data were collected on response times, pod and server utilization, service profile errors, and energy consumption rates. Additionally, the system logged dropped requests and generated cumulative performance metrics. Asynchronous HTTP routines managed concurrent task processing, and replication factor, RF, phases of 2 (38 tags), 5 (95 tags), and 1 (19 tags), which multiply the broadcasted requests from the Ubisense tags in order to simulate a dynamic environment. 
At the conclusion of each experiment iteration, pods were deleted to ensure clean deployments for subsequent runs. Approximately 300,000 log entries per experiment were exported to an on-premise hosted server for further analysis. Energy consumption was continuously monitored using HMC, enabling comparisons of efficiency between small and medium nodes. The experiments included testing category-level and pod-level load balancing to assess their effectiveness under dynamic conditions.


\begin{figure}[ht]
    \centering
    \includegraphics[width=.9\columnwidth]{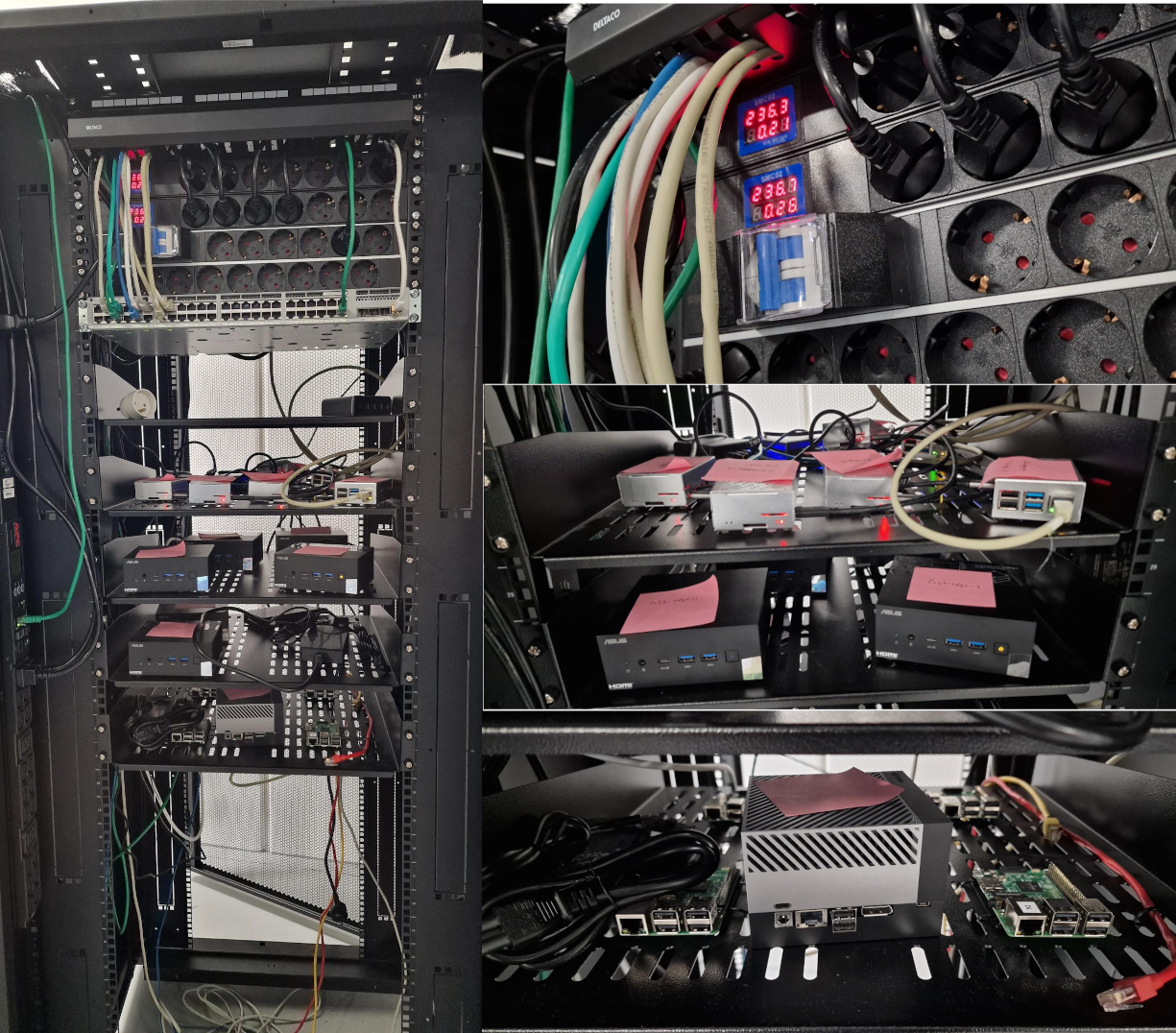}
    \caption{The server rack including the heterogeneous computation nodes.}
    \label{rack}
\end{figure}

\section{Evaluation}

We conducted experiments to compare the three offloading strategies under a mix of workload conditions. Key metrics evaluated include response time, energy consumption, and resource utilization on heterogeneous nodes. The workload was varied in phases to test scalability: for example, we used three phases with increasing numbers of concurrent tasks (RF=2,5,1 corresponding to roughly 38, 95, and 19 requests/sec respectively in each phase) to represent low, peak, and moderate demand periods. Below, we present a comparative analysis of the strategies in terms of the research questions posed.

\subsection{Answer to RQ 1: Impact on System Responsiveness and Energy Efficiency}

The choice of offloading strategy affects both latency and energy per task. Table \ref{tab1} presents a summary of the performance of each strategy, which we will discuss further.

\begin{table}[ht]
\centering
\caption{Comparison of Offloading Strategies}
\begin{scriptsize}
\begin{tabular}{@{}lccc@{}}
\toprule
\textbf{Strategy} & \textbf{Avg. RT (ms)} & \textbf{99th\% (ms)} & \textbf{Energy (mJ)} \\
\midrule
Category-level     & 125 & 300 & 520 \\
Pod-level          & 90  & 180 & 450 \\
\bottomrule
\end{tabular}

\vspace{1mm}
\raggedright
\end{scriptsize}
\label{tab1}
\end{table}

In our measurements, the {\bf category-level strategy} consistently showed the highest response times and the greatest variability. Because this approach sends tasks in equal proportion to small and medium node groups (in our default setup) without fine control, the smaller nodes (Raspberry Pis) often become a bottleneck. For instance, under a high load phase (95 req/s), we observed Raspberry Pi pods hitting 100\% CPU and saturating their throughput, resulting in queuing delays. The PN53 pods, while faster, could not fully compensate since the service-level load balancer kept sending a fixed share of tasks to the Raspberry Pi service. This led to a long-tail latency problem: many tasks on small nodes experienced significantly slower processing (e.g., 2–3× the latency of tasks on medium nodes), pushing the 99th percentile response time to around 300 ms in the worst case (versus ~170–180 ms for the other strategies). The uneven distribution is evidenced by cluster utilization metrics: some {\em small} pods were overloaded while {\em medium} pods had idle capacity at times, due to the static round-robin assignment. Energy efficiency was also suboptimal, although Raspberry Pi devices draw less power (on the order of 3–5 W) compared to PN53 nodes (15–25 W range), their slower processing meant energy per task was higher. Specifically, we found that at high load, a Raspberry Pi consumed ~25\% more energy per inference than a PN53 (normalized by task) because it ran near maximum utilization and sometimes throttled. Thus, the category-level strategy, while functional, tends to waste potential performance and energy efficiency by not leveraging the heterogeneous resources effectively.

The {\bf pod-level strategy} demonstrated clear improvements in both latency and energy metrics. By utilizing real-time metrics to steer tasks to the most capable pods, this strategy reduced the average response time by roughly 25–30\% compared to category-level. As shown in Table 1, the average latency dropped (e.g., from ~125 ms to ~90 ms in our experiments for moderate loads), and the tail latency was significantly lower because fewer tasks were stuck on the slowest resources. The intelligent offloading agent managed to keep all pods busy to an appropriate degree: faster PN53 pods handled a disproportionate share of tasks during peak demand, while Raspberry Pi pods were used sparingly (or for simpler tasks) to avoid slowdowns. This fine-grained load balancing yielded much more uniform pod CPU utilization profiles. In one scenario, during the peak phase (95 req/s), the agent initially routed most tasks to PN53 pods until their CPU usage climbed near a threshold, at which point a small portion of new tasks were diverted to Raspberry Pi pods to prevent queuing on the PN53s. This dynamic redistribution kept response times low on all nodes, i.e., PN53 nodes averaged ~50–60 ms per task, Raspberry Pis ~150 ms, and because relatively few tasks hit the Pis, the overall average remained low.

Consequently, the energy per task also improved: the system completed more tasks per unit time on the energy-efficient high-performance nodes. When tasks were balanced, the PN53 nodes operated at a higher utilization (which is actually more energy-efficient, since their performance per watt increases up to a point), and the Raspberry Pis were not pushed into inefficient high-load regimes. We measured an increase in energy efficiency by ~15\% when using pod-level adaptive balancing versus the equal category split, corroborating that smarter distribution can save energy. A related point is that a Raspberry Pi’s wattage is lower, but its energy per task is worse at high load; by combining node types wisely (as pod-level strategy does), we leveraged the PN53’s superior per-task energy when busy, while still utilizing Pis for light tasks, improving overall energy usage. From a quality-of-service perspective, the pod-level strategy maintained stable performance even as workload intensity changed. The agent quickly reacted to phase shifts (e.g., when the request rate dropped from 95 to 19 req/s, it rebalanced load to ensure PN53s were not idle and Pis handled more of the trickle of tasks, which kept energy usage balanced). Therefore, pod-level adaptation significantly enhanced responsiveness and energy efficiency by using the right resource for each task in real time, validating the benefits of a more granular adaptation mechanism.

\subsection{Answer to RQ 2: Scalability Under Increasing Workload Variability}

The strategies were tested under varying loads, but they differ in how well they scale. The category-level strategy struggled as workload variability increased. In our high-intensity phase, the fixed distribution caused certain pods to become overwhelmed, leading to some tasks being dropped or significantly delayed when a Raspberry Pi hit 100\% CPU for sustained periods (we logged a small number of dropped requests in the category-level runs, whereas pod-level runs had essentially zero drops). Moreover, Kubernetes’ default service load balancer (used at the category level) exhibited performance degradation across sequential experiments. We observed an internal state buildup that led to slower routing decisions, likely due to K3s' implementation of service IP tables. This meant that in back-to-back high-load tests, the category-level approach got worse over time (higher latencies in later runs), indicating poor scalability in long-running scenarios. In contrast, the pod-level strategy scaled smoothly: it maintained stable performance from 10\% to 100\% workload levels by dynamically utilizing all available pods and nodes.

Because it could directly activate more pods (we manually allowed it to scale up to 7 pods per category when needed) and distribute load evenly, the system throughput scaled nearly linearly with added load until the hardware limits were reached. We found that at 95 req/s, the pod-level strategy kept the cluster at about 85\% combined CPU utilization (all PN53 and Pi cores busy) with minimal queuing, effectively using the full capacity. When we increased to an even higher load (beyond our planned scenario), the agent started to drop a few requests only once every pod was saturated, which is expected. This highlights that fine-grained control extends the scalability of the system: it can handle more workload before hitting a bottleneck, compared to coarse strategies.
Thus, with respect to RQ2, the pod-level strategy exhibit strong scalability in handling increased and variable workloads. Category-level, lacking this adaptive finesse, falls short in scaling efficiency, it might require significantly over-provisioning resources to achieve the same performance targets, which is not ideal.

\subsection{Answer to RQ 3: Architectural/Orchestration Considerations for Intelligent Adaptation}

Our findings underline several important techniques necessary to implement runtime adaptation (RQ3). Firstly, high-frequency monitoring and feedback are crucial; the superior performance of pod-level was enabled by our system’s ability to gather metrics at 5Hz and react immediately, something not feasible with standard 15s metrics scraping. This was key to maintaining responsiveness; slower feedback would result in outdated decisions.

Secondly, a robust integration with the orchestration layer (Kubernetes) is needed. Pod-level strategy benefited from the agent’s direct control over pod scheduling and scaling. For example, without the ability to programmatically start/stop pods or query pod IPs on demand, the pod-level strategy could not function. We had to implement custom controllers (via the Kubernetes API) to override default load balancing and to perform on-the-fly scaling. This suggests that future systems should expose hooks for fine-grained control (perhaps via Kubernetes scheduler extenders or service mesh techniques) to implement intelligent offloading.

Thirdly, parallel and asynchronous execution within the adaptation logic is essential. Our offloading agent had to handle dozens of tasks per second, each requiring a decision and a network call; using asynchronous coroutines ensured minimal overhead and allowed the monitor/analyzer to run concurrently with task dispatch. This architecture prevented the agent from becoming a bottleneck and is an important orchestration pattern for real-time adaptive systems.

Additionally, employing a message broker (MQTT) to decouple sensor data ingestion from decision-making proved effective, as it smoothed out bursts and provided a buffer so that no data was lost even if the agent was momentarily busy. 


Implementing intelligent, runtime offloading adaptation requires: {\em i)} an architecture that can monitor and act within seconds or less, {\em ii)} tight coupling with cloud-edge orchestration tools to deploy and migrate workloads as decisions dictate, and {\em iii)} Decision algorithms that can utilize the rich data and act on long-term objectives. Our experiments illustrates how these techniques, working in concert, answer RQ3 by providing a blueprint for deploying self-adaptive offloading in real-world, heterogeneous computing environments.

\subsection{Discussion}
The results above underscore the trade-offs and benefits of each approach. A simple category-level strategy may suffice in homogeneous or low-variability settings, but is outclassed by more adaptive strategies in mixed environments. The pod-level approach offers a compelling improvement with relatively low complexity, essentially adding a smarter overlay over Kubernetes scheduling.
One key insight is that granularity of control is a decisive factor: the more granular and timely the adaptation (down to each task and pod), the better the system can perform, as long as the overhead of that control is kept low. We also note that energy efficiency and performance are sometimes at odds, but a well-designed strategy can manage the trade-off dynamically. 
Regarding scalability, our experiments were on a mid-size edge cluster; the positive results suggest that these strategies would scale to larger clusters as well, though one must consider the computational overhead in the managing system (our agent) as the number of nodes/pods grows. Techniques like hierarchical load balancing (first decide which edge site, then which node, then which pod) or federated learning 
might be needed for very large deployments to keep the decision latency small. These are avenues for future exploration. Overall, the comparative evaluation validates that self-adaptive offloading strategies, particularly those with fine-grained control, substantially improve system responsiveness and energy usage (RQ1), and that they scale effectively with dynamic workloads (RQ2) when underpinned by an appropriate architecture (MAPE-K with high-frequency monitoring and tight orchestration integration, addressing RQ3). 

\section{Lessons Learned}

The experiments and results provided key insights into the design and operation of the intelligent offloading system. These lessons are valuable for refining the current system and guiding future implementations in similar heterogeneous environments.

\begin{itemize}
    \item {\bf Integration of Predictive Models with Real-time Systems.} The deployment of the LSTM-based neural network for mobility behavior predictions highlighted the potential of predictive AI in enhancing robotic operations. By anticipating human movement patterns, the system enabled robots to optimize routing and improve safety in mixed environments. However, as this paper focused on, the real-time requirements of robotics necessitate minimizing computation time and ensuring low-latency responses, which remain key challenges when deploying complex AI models on edge devices.
 \item{\bf Human-Centric System Design.} The inclusion of predictive mobility models showcased the value of designing systems that prioritize human safety and efficiency. Anticipating human actions allows robots to make proactive adjustments, minimizing disruptions and improving the overall workflow. This underscores the importance of human-centric AI in shared environments.
   \item {\bf Importance of Granular Load Balancing.} Pod-level load balancing proved critical for optimizing resource utilization and reducing response times. The ability to assign tasks directly to individual pods based on real-time metrics provided significantly better outcomes than category-level approaches. This reinforces the necessity of incorporating fine-grained control mechanisms into the managing system for handling dynamic workloads.
  \item {\bf Energy-Performance Trade-offs.} The contrast between small and medium nodes highlights the trade-offs between energy efficiency and performance. 
  Balancing these trade-offs through adaptive task allocation is essential for achieving overall system efficiency.
    \item {\bf Impact of High-Frequency Monitoring.} The use of a Rust-based metrics server with high-frequency data collection proved effective in enabling real-time decision-making without overloading the cluster. This demonstrates the importance of lightweight, efficient monitoring solutions in maintaining system responsiveness.
  \item {\bf Scalability and Flexibility.} The system’s ability to scale dynamically by adapting to real-time conditions indicates its potential for deployment in more complex scenarios. However, scalability testing in environments with a greater variety of nodes and more diverse workloads is necessary.
  \item {\bf Value of Heterogeneous Infrastructure.} The experiments highlighted the benefits of using a heterogeneous computing infrastructure, combining low-power and high-capacity nodes. This diversity allows the system to handle a wide range of workloads, from lightweight to compute-intensive tasks, while optimizing for energy efficiency and performance.
  \item {\bf Adaptation Logic as a Scalable Abstraction.}
The case study shows that the MAPE-K feedback loop, when coupled with custom controllers, provided a flexible abstraction for coordinating heterogeneous nodes. This indicates that adaptation logic itself can serve as a scalable architectural building block, decoupling monitoring and decision-making from the specifics of the underlying infrastructure. Future deployments can extend this logic to multi-site or federated edge clusters without redesigning the entire system.

  \item {\bf Data Flow Integration as a Hidden Bottleneck.}
While the primary focus was on computation offloading and adaptation, the experiments revealed that the overall performance of the system also relies heavily on efficient data flow. This flow moves from UWB sensors to MQTT, then through the offloading agent, and finally to the neural network and robot fleet manager. Even with optimized offloading, delays or congestion in these data pipelines can significantly impact performance. This highlights the importance of treating data flow orchestration as equally critical as computation scheduling in real-time robotics environments to achieve system-wide efficiency.
  \item {\bf Potential for Broader Industrial Applications.} The results suggest that the proposed architecture is applicable beyond the tested use case, particularly in other industries requiring dynamic, real-time task management. Expanding the system to include additional predictive models or integrating domain-specific adaptations could further enhance its utility.
\end{itemize}




\section{Conclusion }

This paper presents a MAPE-K-based self-adaptive architecture approach to address the challenges of dynamic task allocation, real-time adaptability, and resource optimization for compute-intensive environments. The approach provides distributed edge computing for predictive AI tasks, enabling intelligent offloading and proactive decision-making. Experiments conducted in a real-world industrial assembly line with humans and robots demonstrated the architecture’s effectiveness in improving response times, optimizing energy consumption, and enhancing scalability. The findings provide insights into balancing performance, energy efficiency, and adaptability in AI-driven environments. They highlight the importance of adaptation capabilities and real-time metrics in managing workloads across heterogeneous infrastructures. Future work will aim to scale the system for larger and more complex environments, refine predictive models to adapt to diverse human-robot interactions, and explore additional quality attributes such as reliability and fault tolerance. These advancements will further establish the proposed architecture as a foundation for intelligent systems in practice.

\section*{Acknowledgment}
This work is supported by the Energi Fyn Development Fund
(Udviklingsfond) 2023, for the project RobotArch.

\bibliographystyle{IEEEtran}
\bibliography{bib}

\end{document}